\documentclass[11pt]{article}

\usepackage[final]{acl}

\usepackage{arydshln}
\usepackage{times}
\usepackage{latexsym}

\usepackage[table,xcdraw]{xcolor}
\usepackage[T1]{fontenc}

\usepackage[utf8]{inputenc}

\usepackage{microtype}
\usepackage{inconsolata}
\usepackage{graphicx}
\usepackage{subcaption}
\usepackage{hyperref}
\usepackage{array} 
\usepackage{multirow} 
\usepackage{algorithm}
\usepackage{algpseudocode} 
\usepackage{booktabs}
\usepackage{arydshln}
\usepackage{amsmath}
\usepackage{amssymb}
\usepackage{tabularx}
\usepackage{enumitem}
\usepackage{pifont}
\usepackage{algorithm}
\usepackage{algpseudocode}
\usepackage{setspace}
\algrenewcommand\alglinenumber[1]{\tiny #1:}
\makeatletter
\renewcommand{\ALG@beginalgorithmic}{\small}
\makeatother

\title{Improving Synthetic Data Generation for Argument Mining via Adversarial Reinforcement Learning}

\author{
Zhijun Zhang\textsuperscript{1},
Qianlong Wang\textsuperscript{2,\hyperlink{corr}{\ensuremath{\ast}}},
Keyang Ding\textsuperscript{3},
Genan Dai\textsuperscript{2}, \\
\textbf{Bowen Zhang\textsuperscript{2}},
\textbf{Bin Liang\textsuperscript{4}},
\textbf{Ruifeng Xu\textsuperscript{3,5}},
\textbf{Yongsheng Liang\textsuperscript{1,2,3}}
\\
\textsuperscript{1}College of Applied Science, Shenzhen University, Shenzhen, China \\
\textsuperscript{2}School of Artificial Intelligence, Shenzhen Technology University, Shenzhen, China \\
\textsuperscript{3}Harbin Institute of Technology (Shenzhen), Shenzhen, China \\
\textsuperscript{4}The Chinese University of Hong Kong, Hong Kong, China \\
\textsuperscript{5}Pengcheng Laboratory, Shenzhen, China \\
\texttt{2510263003@mails.szu.edu.cn, wangqianlong@sztu.edu.cn}
}

\begin{document}
\maketitle

\begingroup
\renewcommand{\thefootnote}{}
\footnotetext{%
  \hypertarget{corr}{}%
  \textsuperscript{\ensuremath{\ast}} Corresponding author.%
}
\endgroup

\begin{abstract}
Argument Mining (AM) is fundamentally constrained by the scarcity of high-quality structure-annotated datasets.
While LLMs have shown promise in synthetic data generation, producing synthetic AM data that is both structurally accurate and sufficiently diverse remains a challenging problem.
To address this problem, we revisit synthetic data generation for AM from a new perspective and propose a novel \textit{adversarial reinforcement learning framework for data synthesis}.
The proposed framework jointly optimizes the generator and the discriminator in an adversarial loop, in which the generator produces structured AM instances, and the discriminator provides learning signals by distinguishing real data from synthetic candidates. This enables the generator to progressively improve both the structural accuracy of generated argument data while maintaining diversity through adversarial feedback.
Extensive experiments demonstrate that the proposed framework consistently improves AM performance on three benchmark datasets in both full-data and low-resource settings, validating its effectiveness and scalability.
\end{abstract}

\section{Introduction}
Argument Mining (AM) aims to identify argument structures in natural language text automatically\footnote{The task definition of AM can be found in Appendix \ref{appendix:Task-Formulation}.}. Specifically, it involves detecting argumentative components, such as claims and premises, and identifying the relations among them, including support and attack relations \citep{lippi2015context}.
As one of the core tasks in computational argumentation, AM is highly significant for downstream applications such as argument retrieval, opinion analysis, and argument generation \citep{wachsmuth2017building,lawrence2019argument,chen2024exploring}.

However, the development of end-to-end AM models has long been constrained by data scarcity \citep{dutta2022can,morio2022end}. 
This is because AM requires not only accurately determining whether a text span is argumentative, but also performing fine-grained type classification of components and relation mapping at the structural level. 
Such high complexity makes the manual construction of high-quality AM corpora usually costly \citep{das2018developing}.
As a result, existing benchmark datasets for AM are typically small in scale despite their research value, which limits the training of robust and high-performing AM models.
For instance, the AAEC dataset \cite{stab2017parsing} contains only 289 training essays.

In recent years, the development of large language models (LLMs) has provided new possibilities for alleviating data scarcity in AM. 
Although LLM-based synthetic data generation has shown remarkable effectiveness in improving the performance of various NLP tasks \citep{tan2024large,dai2025auggpt}, generating synthetic data for AM remains highly challenging. 
This is because high-quality AM samples should not only be natural and fluent at the linguistic level, but also preserve structurally valid argumentative relations and component organization \citep{persing2016end}. 
Moreover, the AM samples should maintain sufficient diversity in both topics and argument structures.
Recent studies have begun exploring this direction. In particular, \citeauthor{bao2025exploring} leveraged prompt engineering with GPT-4o to enhance both the quality and diversity of synthetic AM data.
Despite promising gains, this approach still relies heavily on costly closed-source LLMs and treats quality and diversity in separate generation pipelines, which makes it difficult to produce synthetic samples that are both structurally reliable and sufficiently diverse.

Building upon these limitations, we argue that synthetic data generation for AM should jointly optimize data quality and diversity within a unified framework. 
A straightforward solution is to employ open-source LLMs with stochastic sampling strategies to increase output diversity while generating high-quality synthetic data. However, sampling and generating without explicit supervision signals often lead to noisy instances. 
In this context, reinforcement learning may be considered as a natural mechanism to incorporate some signals and optimize the generation process toward both quality and diversity, thereby encouraging outputs that more closely resemble real data.


Motivated by this, we approach synthetic data generation for AM from a new perspective and propose a novel \textit{Adversarial Reinforcement Learning framework for Data Synthesis}.
Our key idea is to jointly train two LLMs in an adversarial manner: the generator and the discriminator, where reinforcement learning is used to propagate the discriminator’s feedback as learning signals for the generator.
Specifically, the generator first samples some synthetic AM data conditioned on gold-annotated data. The discriminator is then presented with a candidate set containing one annotated sample and multiple synthetic samples, and is required to identify the real instance.
The discriminator’s decisions are converted into reward signals and further combined with rule-based constraints to form the final training signal. Here, we optimize the generator using Group Relative Policy Optimization (GRPO) \citep{shao2024deepseekmath}, enabling it to produce increasingly realistic and diverse synthetic data.
Meanwhile, the generator is encouraged to produce challenging samples that the discriminator cannot reliably distinguish from real annotated data.
Subsequently, the improved generator produces more diverse and harder-to-detect synthetic samples, which are then used to further refine the discriminator’s ability.
Here, the discriminator is trained to maximize its identification accuracy.
Through this alternating adversarial optimization process, both LLMs improve iteratively.
As a result, the generator progressively learns to produce synthetic AM data with higher linguistic naturalness, more accurate argument structures, and greater diversity.

We conduct comprehensive experiments on three widely used AM datasets. Empirical results show that augmenting the original training data with synthetic samples generated by our framework (implemented with Qwen3-8B) consistently yields substantial improvements over strong AM baselines across both full-data and low-resource settings.
Notably, our framework even achieves performance surpassing data generated by stronger proprietary systems such as GPT-4o. Further analysis confirms that the gains stem from both improved linguistic naturalness and greater structural diversity of the generated data.

\section{Related Work}

\subsection{Argument Mining}
AM aims to identify argumentative content and its internal structure in unstructured text \citep{lawrence2019argument}. 
Argument structures involve argument components, component types, and logical relations between components. 
Thus, existing studies decompose the task into several more tractable subtasks, including argument component segmentation and classification, which identify component spans in text \citep{moens2007automatic,stab2017parsing,lippi2015context}, and argumentative relation identification, which determines whether semantic relations exist between components \citep{cocarascu2017identifying,jo2021classifying}. 

Such staged modeling approaches are susceptible to error propagation. Consequently, growing attention has recently been devoted to joint modeling and end-to-end AM methods \citep{persing2016end,eger2017neural}.
Early methods mainly drew on natural language parsing techniques, such as sequence labeling \citep{wang2021progressive,ye2021end} or logical inference problems \citep{bao2021neural,morio2022end}. With the development of LLMs, AM has gradually been reformulated as a seq2seq generation task \citep{kawarada2024argument,sun2024discourse,bao2025uniasa}. 

However, AM models are still constrained by the scarcity of annotated corpora that limits further performance improvements. This is mainly due to the inherent complexity of argument structures, making data annotation highly costly. 

\begin{figure*}[t]
  \includegraphics[width=0.95\textwidth]{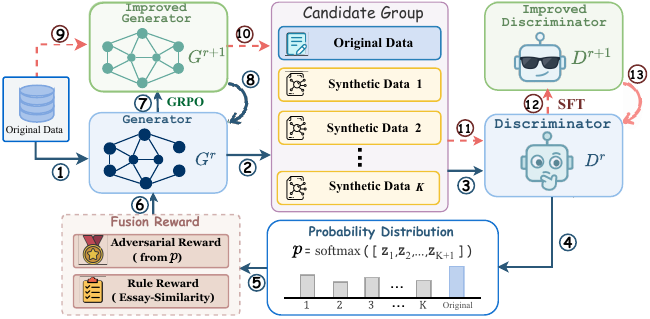}
\vspace{-3mm}
  \caption{Overview of the proposed framework. The framework consists of two iterative stages: generator optimization (\normalsize{\textcircled{\small{1}}}--\normalsize{\textcircled{\small{8}}}) and discriminator refinement (\normalsize{\textcircled{\small{9}}}--\normalsize{\textcircled{\small{13}}}). 
  Through alternating adversarial training, the generator and discriminator progressively improve together.}
\vspace{-4mm}
\label{fig:framework}
\end{figure*}

\subsection{Synthetic Data Generation}
Synthetic data generation has become an important approach for data augmentation in NLP, motivating researchers to leverage LLMs to construct task-specific training samples
\citep{bao2023synthetic,wang2023retrieval,guo2024generative,havrilla2024surveying}. 
Compared with general generations, synthetic data generation for structured-output tasks needs to ensure both text quality and the correctness of label structures. Thus, task-specific constraints and structure-preserving mechanisms are needed to control the quality of synthetic samples \citep{josifoski2023exploiting,zhang2024llm}. 

Recent studies have explored generating AM data with GPT-4o from both quality and diversity perspectives, showing that synthetic data can improve AM model performance \citep{bao2025exploring}.
However, existing GPT-4o-based approaches are costly and rely heavily on prompt engineering, making data quality and diversity difficult to control without model adaptation. 

In recent years, reinforcement learning and preference optimization have also been increasingly applied to synthetic data generation, where prior studies optimize the data generation process using preference signals or downstream feedback \citep{wang2024lrhp,hou2025private,zweiger2026self,wang2026mro,wang2026msrl}.
However, these methods mainly target general data generation scenarios and are not specifically designed to maintain consistency between argumentative text and structured AM annotations. This work addresses this research gap by exploring the application of reinforcement learning to AM data synthesis.


\section{Framework}
\subsection{Overview}
As illustrated in Figure \ref{fig:framework}, our framework consists of two iterative stages: generator optimization and discriminator refinement. 
The generator $G_{\theta}$ first produces multiple synthetic AM samples based on an annotated instance, and is optimized using reward signals derived from the discriminator’s feedback. 
The discriminator $D_{\phi}$ is then further trained to distinguish annotated samples from newly generated synthetic samples. 
These two stages are alternately repeated for several rounds, enabling both models to progressively improve through adversarial interaction.

\subsection{Generator Optimization Stage}
Before adversarial training, we initialize the generator to provide a basic ability to produce valid AM synthetic samples. Let the annotated AM dataset be $\mathcal{D}=\{z_i \}_{i=1}^{N}$, where each sample $z_i=(x_i,y_i)$ consists of an input text $x_i$ and its corresponding structured argument annotation $y_i$.

\paragraph{Generator initialization.}
Given an annotated sample $z$, we serialize it into an instruction-style prompt $u$. The target output $v$ is a synthetic AM sample that follows the same annotation schema while differing from the annotated sample in both textual content and argument structure. The generator is initialized through supervised fine-tuning (SFT) with the following objective:
\begin{equation}
\mathcal{L}_{\mathrm{SFT}}(\theta)=-\mathbb{E}_{(u,v)\sim z,z \sim \mathcal{D}}\log p_{\theta}(v \mid u)\label{eq:sft}
\end{equation}
Here, $u$ denotes the prompt constructed from the real AM sample $x$, and $v$ denotes the corresponding target synthetic sample\footnote{Initially generated using Qwen3-8B with a detailed prompt, followed by basic error filtering.}. During this stage, the generator learns to produce complete text-annotation pairs that follow the target AM schema.

\paragraph{Reinforcement learning.}
After initialization, the generator is further optimized through adversarial reinforcement learning using feedback from the discriminator.
Specifically, for each annotated sample $z$, the generator produces $K$ synthetic samples. 
We then construct a shuffled candidate set $C$ consisting of one annotated sample and $K$ synthetic samples. 
Let $y_{\mathrm{position}}$ represent the position of the annotated sample. 
To encourage the generator to produce high-quality samples that can confuse the discriminator, the discriminator reward is defined:
\begin{equation}
R_{\mathrm{dis}}=\min\left(-\log\left(p_{\phi}(y_{\mathrm{position}}\mid C)+\epsilon\right),c\right)
\label{eq:disc_reward}
\end{equation}
where $\epsilon$ is a small constant for numerical stability, and $c$ is the reward clipping threshold. 
Here, a larger reward indicates that the discriminator finds it more difficult to identify the annotated sample, suggesting that the generated samples are closer to high-quality real AM data. The clipping operation prevents excessively large rewards that may destabilize training.

In preliminary experiments, we find that the generator sometimes makes only shallow modifications to the source sample, or even directly reuses expressions from the original data. To alleviate this issue, we introduce a source similarity penalty that discourages generated samples from being overly similar to the source sample. Specifically, we compute the ROUGE-L F1 similarity between the synthetic text and the source text. 
Here, if the similarity is below a threshold $\tau$, no penalty is applied. Otherwise, the penalty increases with similarity and is bounded within $[-1,0]$. Let $R_{\mathrm{sim}}^{k}$ denote the similarity penalty for the $k$-th synthetic sample. The final reward is defined as:
\begin{equation}
R_k= R_{\mathrm{dis}}+\lambda_1 \times R_{\mathrm{sim}}^{k}
\label{eq:final_reward}
\end{equation}
Here, $\lambda_1$ is the weight of the similarity penalty. This reward encourages the generator to produce synthetic samples that are more realistic and difficult to distinguish, while the similarity penalty explicitly promotes sample novelty and diversity.

After obtaining the final rewards, we optimize the generator using GRPO, which estimates the relative advantage of each synthetic sample based on the rewards of all outputs within the same group. Let $A_k$ denote the advantage of the $k$-th synthetic sample, and let $\rho_{k,t}(\theta)$ denote the probability ratio between the current policy and the old policy at the $t$-step token. The optimization objective is as follows:
\begin{align}
\mathcal{L}_{\mathrm{GRPO}}(\theta)&=-\mathbb{E}_{\{\hat{z_k}\}_{k=1}^{K} \overset{G_{\theta}}{\sim} z, z \sim \mathcal{D}}\Big[\min\big(\rho_{k,t}(\theta)A_k, \notag\\
& \hspace{-2.5em}
\operatorname{clip}\big(\rho_{k,t}(\theta), 1-\epsilon_c,1 + \epsilon_c\big)A_k\big)\Big]
\label{eq:grpo_loss}
\end{align}
This objective increases the generation probability of high-reward outputs, thereby encouraging the generator to produce more realistic, diverse, and structurally valid AM synthetic data.

\subsection{Discriminator Refinement Stage}
This stage focuses on iteratively improving the discriminator’s ability to distinguish annotated AM samples from increasingly high-quality synthetic ones generated by the evolving generator.

\paragraph{Discriminator Initialization.}
For the discriminator $D_{\phi}$, each training instance is a candidate set consisting of one annotated sample and $K$ synthetic samples. 
The synthetic samples may contain various types of noise, including unnatural expressions, misaligned argument spans, or invalid relations.
Given a shuffled candidate set $C$, the discriminator outputs a probability distribution over all positions $\left[p_{\phi}(j\mid C)\right]_{j=0}^{K+1}$, where each value is the likelihood that the corresponding sample is the real one. Here, the index of the annotated sample is used as the supervision signal. The equation is as follows:
\begin{equation}
\mathcal{L}_{\mathrm{SFT}}(\phi)=-\log p_{\phi}(y_{position} \mid C)
\label{eq:disc_output}
\end{equation}
Based on this, the discriminator is trained as a multi-class classifier to identify the real sample, thereby gradually learning to detect quality issues in textual realization and structural annotations.

\paragraph{Supervised Refinement.}
After each generator update, the improved generator produces more realistic and challenging synthetic samples. These newly generated samples are then incorporated into the training data to further fine-tune the discriminator using the supervised objective in Eq.~\ref{eq:disc_output}.
As a result, the discriminator progressively becomes more sensitive to subtle structural inconsistencies and annotation errors, leading to stronger discrimination ability against high-quality synthetic samples produced by the evolving generator.

\subsection{Adversarial Evolution}
Our framework is optimized through iterative adversarial evolution. In the $r$-th evolution round, the current generator $G_{\theta}^{r}$ first produces multiple synthetic AM samples conditioned on an annotated sample. A candidate set is then constructed, and the discriminator identifies the real sample among the candidates.
The discriminator output is converted into a reward signal and further combined with the source similarity penalty. The resulting reward is used to optimize the generator via GRPO, producing an updated generator $G_{\theta}^{r+1}$.
These newly generated data are subsequently used to further refine the discriminator $D_{\phi}^{r+1}$, improving its ability to distinguish real samples from synthetic samples. Through this alternating optimization process, the generator and discriminator evolve together progressively.
The detailed pseudocode of this process is provided in Appendix \ref{appendix:Algorithm}.

\section{Experiments}

\subsection{Datasets}
We conduct experiments on three widely used AM datasets: AbstRCT \citep{mayer2020transformer}, AAEC \citep{stab2017parsing}, and CDCP \citep{park2018corpus}. 
Each dataset contains argumentative texts and complete manually annotated results. Additional details about the datasets' construction and statistics are provided in Appendix \ref{appendix:datasets}. Following \citeauthor{bao2025exploring}, our experiments are performed under two training data settings: using 100\% of the original training data and a low-resource setting with only 5\% of the training data.

\subsection{Implementation Details}
In this work, we use Qwen3-8B as the base model for both the generator and the discriminator. 
We set the number of adversarial evolution rounds to $r$=3, and use $K$=3, and $\lambda_1$=0.5 in all experiments. 
Following \citeauthor{bao2025exploring}, we limit the scale of the generated synthetic data to twice the size of the original training data used under the selected setting. All experiments were conducted on an NVIDIA H20 GPU.
For all experiments\footnote{Code is available at \url{https://github.com/ZhangZJ-GH/Synthetic-Data-for-Argument-Mining}.}, we report the average results obtained over five different random seeds. 

We use three fine-grained F1 scores from prior work as evaluation metrics: $\mathrm{\textbf{F1}}_{\textbf{span}}$ measures the matching accuracy of component spans, $\mathrm{\textbf{F1}}_{\textbf{aci}}$ evaluates the accuracy of component type classification, and $\mathrm{\textbf{F1}}_{\textbf{ari}}$ assesses the mapping accuracy of argumentative relations.

\subsection{AM Models}
To examine the effectiveness of synthetic data under different model settings, we adopt three AM models. (1) \textbf{ST} \citep{morio2022end} represents parsing-based AM methods and explicitly performs argument component identification, component type classification, and argumentative relation prediction. (2) \textbf{UniASA} \citep{bao2025uniasa} is a task-specific generative AM method by reformulating AM as a seq2seq generation task, and we adopt its single-view version to reduce computational cost. (3) \textbf{Directly-SFT} serves as an LLM-native generative baseline, where AM data are constructed as instruction-tuning samples and LLMs are directly fine-tuned to generate argument structures. 
Here, we use Qwen3-8B \cite{yang2025qwen3} as the base model. 

\subsection{Compared Methods}
We compare our framework with the following approaches: (1) \textbf{EDA} (Easy Data Augmentation), which applies synonym replacement to original argumentative texts;
(2) \textbf{FTGA} (Few-shot Text Generation and Auto-annotation), which uses GPT-4o to generate new texts and then labels them with an AM model; 
(3) \textbf{JTLS} (Joint Text and Label Synthesis), which prompts GPT-4o to generate both texts and AM annotations.
(4) \textbf{QOS} (Quality-Oriented Synthesis), which rewrites original samples while preserving their argument structures. 
(5) \textbf{DOS} (Diversity-Oriented Synthesis), which generates diverse samples by varying topics and argumentation patterns.
(6) \textbf{QOS+DOS} \citep{bao2025exploring}, which combines 25\% QOS data and 75\% DOS data, as this combination has been shown to achieve the best performance among different mixing ratios; and (7) \textbf{SFTSyn} (Supervised Fine-Tuned Synthesis), where LLMs are fine-tuned to directly generate synthetic text-annotation pairs. Here, we use Qwen3-8B as the LLM.
More implementation details are provided in Appendix \ref{appendix:method}.

\begin{table*}[t!]
    \normalsize 
    \centering
    \renewcommand{\arraystretch}{1.05}
    \resizebox{\linewidth}{!}{
         \begin{tabular}{lllcccccccccccc}
  \toprule[1.5pt]
   &  &  & \multicolumn{4}{c}{\textbf{AAEC}} & \multicolumn{4}{c}{\textbf{AbstRCT}} & \multicolumn{4}{c}{\textbf{CDCP}} \\
  \cmidrule[\heavyrulewidth](lr){4-7} \cmidrule[\heavyrulewidth](lr){8-11}  \cmidrule[\heavyrulewidth](lr){12-15}
  \multirow{-2}{*}{\textbf{AM Model}} & \multirow{-2}{*}{\textbf{Setting}} & \multirow{-2}{*}{\textbf{Method}} & {\textbf{$\mathrm{F1}_{span}$}} & {\textbf{$\mathrm{F1}_{aci}$}} & {\textbf{$\mathrm{F1}_{ari}$}} & {\textbf{Avg.}} & {\textbf{$\mathrm{F1}_{span}$}} & {\textbf{$\mathrm{F1}_{aci}$}} & {\textbf{$\mathrm{F1}_{ari}$}} & {\textbf{Avg.}} & {\textbf{$\mathrm{F1}_{span}$}} & {\textbf{$\mathrm{F1}_{aci}$}} & {\textbf{$\mathrm{F1}_{ari}$}} & {\textbf{Avg.}} \\
  \toprule[0.8pt]
   &  & Origin & 84.70 & 75.88 & 54.16 & 71.58 & 69.01 & 63.34 & 38.43 & 56.93 & 82.12 & 68.60 & 31.07 & 60.60 \\\cdashline{3-15}
   &  & EDA & 85.20 & 76.62 & 54.18 & 72.00 & 69.74 & 63.91 & 40.07 & 57.91 & 82.37 & 68.64 & 30.97 & 60.66 \\
   &  & FTGA & 85.71 & 77.00 & 54.60 & 72.44 & 70.47 & 63.98 & 40.16 & 58.20 & 81.89 & 68.54 & 32.95 & 61.13 \\
   &  & JTLS & 85.60 & 76.35 & 54.56 & 72.17 & 70.15 & 64.32 & 40.64 & 58.37 & 82.22 & 68.69 & 30.95 & 60.62 \\
   &  & QOS & 86.19 & 76.57 & 56.75 & 73.17 & 72.53 & 66.08 & 39.49 & 59.37 & 82.72 & 68.62 & 33.75 & 61.70 \\
   &  & DOS & 85.63 & 76.94 & 55.98 & 72.85 & 71.90 & 65.41 & 38.43 & 58.58 & 83.04 & 67.48 & 33.17 & 61.23 \\
   &  & QOS+DOS & 85.98 & 77.08 & 56.99 & 73.35 & 73.19 & 67.12 & 40.19 & 60.17 & 83.72 & 69.06 & 34.66 & 62.48 \\   \cdashline{3-15}
   &  & SFTSyn & 84.11 & 75.07 & 52.33 & 70.50 & 68.14 & 62.03 & 33.84 & 54.67 & 81.00 & 66.52 & 29.59 & 59.04 \\
   & \multirow{-9}{*}{100\%} & \textbf{Ours} & \textbf{88.16} & \textbf{79.63} & \textbf{59.87} & \textbf{75.89} & \textbf{74.03} & \textbf{69.45} & \textbf{41.74} & \textbf{61.74} & \textbf{85.32} & \textbf{70.03} & \textbf{38.16} & \textbf{64.50} \\
  \cline{2-15}
   &  & Origin & 68.24 & 50.42 & 16.96 & 45.21 & 57.17 & 50.13 & 21.50 & 42.93 & 76.84 & 49.03 & 3.40 & 43.09 \\\cdashline{3-15}
   &  & EDA & 67.05 & 48.49 & 16.48 & 44.01 & 58.00 & 51.16 & 21.52 & 43.56 & 76.54 & 47.80 & 3.64 & 42.66 \\
   &  & FTGA & 69.50 & 52.93 & 20.88 & 47.77 & 56.61 & 50.30 & 22.95 & 43.29 & 76.40 & 49.28 & 4.44 & 43.37 \\
   &  & JTLS & 71.16 & 52.51 & 20.23 & 47.97 & 59.12 & 50.32 & 22.20 & 43.88 & 77.61 & 48.96 & 3.20 & 43.26 \\
   &  & QOS & 70.84 & 56.52 & 23.04 & 50.13 & 61.87 & 54.95 & 28.21 & 48.34 & 77.42 & 51.55 & 7.03 & 45.33 \\
   &  & DOS & 71.62 & 54.85 & 23.12 & 49.86 & 60.58 & 53.01 & 23.24 & 45.61 & 76.59 & 52.22 & 6.93 & 45.25 \\
   &  & QOS+DOS & 72.10 & 56.02 & 23.45 & 50.52 & 63.04 & 56.22 & 28.18 & 49.15 & 77.60 & 54.80 & 7.87 & 46.76 \\  \cdashline{3-15}
   &  & SFTSyn & 68.35 & 48.11 & 15.55 & 44.00 & 57.14 & 49.93 & 19.71 & 42.26 & 75.01 & 48.23 & 4.93 & 42.72 \\
  \multirow{-18}{*}{ST} & \multirow{-9}{*}{5\%} & \textbf{Ours} & \textbf{74.74} & \textbf{58.19} & \textbf{26.88} & \textbf{53.27} & \textbf{65.53} & \textbf{58.30} & \textbf{29.39} & \textbf{51.07} & \textbf{79.10} & \textbf{55.24} & \textbf{11.93} & \textbf{48.76} \\
  \toprule[0.8pt]
   &  & Origin & 86.72 & 77.13 & 56.21 & 73.35 & 74.67 & 66.05 & 34.01 & 58.24 & 81.85 & 67.08 & 25.57 & 58.17 \\\cdashline{3-15}
   &  & EDA & 87.02 & 77.69 & 56.95 & 73.89 & 73.64 & 66.34 & 34.30 & 58.09 & 81.60 & 67.08 & 24.58 & 57.75 \\
   &  & FTGA & 87.18 & 77.68 & 56.70 & 73.85 & 74.90 & 66.10 & 35.39 & 58.80 & 82.05 & 67.93 & 24.73 & 58.24 \\
   &  & JTLS & 87.08 & 77.62 & 56.56 & 73.75 & 74.98 & 66.52 & 35.31 & 58.94 & 82.46 & 67.87 & 25.08 & 58.47 \\
   &  & QOS & 87.50 & 79.02 & 57.58 & 74.70 & 75.75 & 67.33 & 36.32 & 59.80 & 82.05 & 66.12 & 25.22 & 57.80 \\
   &  & DOS & 87.49 & 79.41 & 57.61 & 74.84 & 74.06 & 66.95 & 37.67 & 59.56 & 77.90 & 57.89 & 25.28 & 53.69 \\
   &  & QOS+DOS & 88.15 & 78.60 & 58.81 & 75.19 & 75.98 & 68.12 & 39.41 & 61.17 & 82.32 & 68.04 & 27.38 & 59.25 \\   \cdashline{3-15}
   &  & SFTSyn & 85.96 & 75.68 & 53.85 & 71.83 & 73.11 & 65.01 & 31.54 & 56.55 & 80.75 & 65.00 & 24.17 & 56.64 \\
   & \multirow{-9}{*}{100\%} & \textbf{Ours} & \textbf{89.54} & \textbf{80.83} & \textbf{61.24} & \textbf{77.20} & \textbf{77.92} & \textbf{68.45} & \textbf{42.28} & \textbf{62.88} & \textbf{84.10} & \textbf{69.65} & \textbf{30.11} & \textbf{61.29} \\
  \cline{2-15}
   &  & Origin & 54.50 & 34.87 & 5.35 & 31.57 & 57.25 & 44.33 & 12.69 & 38.09 & 71.35 & 32.03 & 2.90 & 35.43 \\\cdashline{3-15}
   &  & EDA & 61.68 & 41.42 & 6.35 & 36.48 & 58.03 & 41.42 & 12.96 & 37.47 & 72.69 & 35.68 & 4.67 & 37.68 \\
   &  & FTGA & 69.85 & 47.81 & 7.98 & 41.88 & 57.87 & 46.18 & 15.31 & 39.79 & 73.48 & 34.33 & 5.31 & 37.71 \\
   &  & JTLS & 58.84 & 39.09 & 6.37 & 34.77 & 57.60 & 43.33 & 16.34 & 39.09 & 73.25 & 37.83 & 3.56 & 38.21 \\
   &  & QOS & 75.92 & 57.33 & 10.15 & 47.80 & 59.31 & 52.00 & 18.44 & 43.25 & 71.60 & 45.32 & 6.98 & 41.30 \\
   &  & DOS & 74.45 & 54.04 & 8.69 & 45.73 & 59.94 & 50.69 & 16.59 & 42.41 & 71.91 & 44.19 & 8.01 & 41.37 \\
   &  & QOS+DOS & 76.69 & 57.19 & 13.12 & 49.00 & 60.32 & 54.22 & 22.80 & 45.78 & 72.83 & \textbf{47.07} & 8.19 & 42.70 \\  \cdashline{3-15}
   &  & SFTSyn & 60.73 & 42.76 & 6.23 & 36.57 & 56.09 & 43.78 & 13.40 & 37.76 & 70.44 & 33.69 & 2.03 & 35.39 \\
  \multirow{-18}{*}{UniASA} & \multirow{-9}{*}{5\%} & \textbf{Ours} & \textbf{77.60} & \textbf{58.26} & \textbf{15.03} & \textbf{50.30} & \textbf{62.64} & \textbf{56.64} & \textbf{26.13} & \textbf{48.47} & \textbf{74.40} & 46.72 & \textbf{11.03} & \textbf{44.05} \\
  \toprule[0.8pt]
   &  & Origin & 78.54 & 66.83 & 37.64 & 61.00 & 73.07 & 65.52 & 47.03 & 61.87 & 77.32 & 59.04 & 21.38 & 52.58 \\\cdashline{3-15}
   &  & EDA & 78.92 & 67.37 & 37.98 & 61.42 & 72.95 & 65.95 & 47.82 & 62.24 & 77.25 & 58.86 & 20.86 & 52.32 \\
   &  & FTGA & 79.28 & 67.51 & 38.52 & 61.77 & 73.76 & 65.86 & 48.37 & 62.66 & 77.55 & 59.18 & 22.05 & 52.93 \\
   &  & JTLS & 79.16 & 67.42 & 38.39 & 61.66 & 73.67 & 66.14 & 48.65 & 62.82 & 77.74 & 59.12 & 21.24 & 52.70 \\
   &  & QOS & 79.83 & 66.63 & 39.41 & 61.96 & 73.49 & 66.27 & 49.30 & 63.02 & 77.14 & 56.70 & 23.40 & 52.41 \\
   &  & DOS & 82.02 & 67.95 & 40.09 & 63.35 & 73.17 & 65.41 & 48.94 & 62.51 & 76.66 & 58.23 & 23.46 & 52.78 \\
   &  & QOS+DOS & 81.99 & 70.13 & 39.51 & 63.88 & 74.01 & 66.74 & 49.21 & 63.32 & 78.15 & 57.65 & 24.58 & 53.46 \\   \cdashline{3-15}
   &  & SFTSyn & 79.03 & 66.35 & 34.43 & 59.94 & 70.63 & 65.05 & 47.41 & 61.03 & 74.37 & 57.83 & 17.44 & 49.88 \\
   & \multirow{-9}{*}{100\%} & \textbf{Ours} & \textbf{84.48} & \textbf{71.10} & \textbf{42.91} & \textbf{66.16} & \textbf{74.51} & \textbf{73.88} & \textbf{52.80} & \textbf{67.06} & \textbf{79.66} & \textbf{59.57} & \textbf{28.03} & \textbf{55.75} \\
  \cline{2-15}
   &  & Origin & 64.76 & 51.26 & 11.03 & 42.35 & 64.94 & 56.82 & 35.17 & 52.31 & 67.92 & 37.95 & 9.28 & 38.38 \\\cdashline{3-15}
   &  & EDA & 65.12 & 50.84 & 10.58 & 42.18 & 65.42 & 56.08 & 35.31 & 52.27 & 68.36 & 38.44 & 9.85 & 38.88 \\
   &  & FTGA & 65.86 & 51.72 & 12.38 & 43.32 & 65.24 & 57.22 & 36.92 & 53.13 & 68.81 & 38.76 & 10.72 & 39.43 \\
   &  & JTLS & 66.28 & 51.62 & 12.14 & 43.35 & 66.04 & 56.56 & 37.14 & 53.25 & 69.15 & 38.62 & 9.96 & 39.24 \\
   &  & QOS & 65.94 & 51.26 & 11.60 & 42.93 & 67.23 & 56.08 & 37.76 & 53.69 & 69.58 & 35.15 & 10.54 & 38.42 \\
   &  & DOS & 66.01 & 51.94 & 12.89 & 43.61 & 66.71 & 55.30 & 36.42 & 52.81 & 69.03 & 36.74 & 11.06 & 38.94 \\
   &  & QOS+DOS & 66.95 & 51.59 & 14.12 & 44.22 & 67.44 & 55.32 & 37.83 & 53.53 & 69.60 & 37.77 & 11.83 & 39.73 \\\cdashline{3-15}
   &  & SFTSyn & 63.94 & 50.88 & 6.48 & 40.43 & 62.87 & 56.11 & 35.58 & 51.52 & 63.17 & 36.37 & 5.76 & 35.10 \\
  \multirow{-18}{*}{Directly-SFT} & \multirow{-9}{*}{5\%} & \textbf{Ours} & \textbf{68.48} & \textbf{52.91} & \textbf{19.07} & \textbf{46.82} & \textbf{68.92} & \textbf{62.47} & \textbf{38.78} & \textbf{56.72} & \textbf{71.97} & \textbf{40.13} & \textbf{13.62} & \textbf{41.91} \\
  \bottomrule[1pt]
\end{tabular}
    }
    \vspace{-2mm}
    \caption{Main experimental results under full (100\%) and low-resource (5\%) training data settings. \textbf{Bold} numbers denote the best performance among all methods on each dataset. “\textbf{Avg.}” is the arithmetic mean of the three F1 scores. The significance test results indicate that all improvements on Avg. are statistically significant with \textit{p} < 0.05.}
    \label{tab:main result}
    \vspace{-4mm}
\end{table*}

\begin{table*}[t!]
\centering
\resizebox{\linewidth}{!}{
\begin{tabular}{llcccccccccccc}
\toprule[1.5pt]
 &  & \multicolumn{4}{c}{\textbf{AAEC}} & \multicolumn{4}{c}{\textbf{AbstRCT}} & \multicolumn{4}{c}{\textbf{CDCP}} \\
\cmidrule[\heavyrulewidth](lr){3-6}
\cmidrule[\heavyrulewidth](lr){7-10}
\cmidrule[\heavyrulewidth](lr){11-14}
\multirow{-2}{*}{\centering\textbf{Setting}} 
& \multirow{-2}{*}{\centering\textbf{Method}} 
& \textbf{$\mathrm{F1}_{span}$} & \textbf{$\mathrm{F1}_{aci}$} & \textbf{$\mathrm{F1}_{ari}$} & \textbf{Avg.}
& \textbf{$\mathrm{F1}_{span}$} & \textbf{$\mathrm{F1}_{aci}$} & \textbf{$\mathrm{F1}_{ari}$} & \textbf{Avg.}
& \textbf{$\mathrm{F1}_{span}$} & \textbf{$\mathrm{F1}_{aci}$} & \textbf{$\mathrm{F1}_{ari}$} & \textbf{Avg.} \\
\hline
 & \textbf{Ours} & \textbf{88.16} & \textbf{79.63} & \textbf{59.87} & \textbf{75.89}
& \textbf{74.03} & \textbf{69.45} & \textbf{41.74} & \textbf{61.74}
& \textbf{85.32} & \textbf{70.03} & \textbf{38.16} & \textbf{64.50} \\
 & \textit{w/o Disc. Refine} & 84.97 & 75.75 & 57.11 & 72.61
& 72.23 & 66.52 & 40.83 & 59.86
& 83.86 & 67.50 & 34.30 & 61.89 \\
\multirow{-3}{*}{\centering 100\%} & \textit{w/o Sim. Penalty} & 84.44 & 78.85 & 59.32 & 74.20
& 73.86 & 68.93 & 41.25 & 61.35
& 84.84 & 68.85 & 37.10 & 63.60 \\
\cline{1-14}
 & \textbf{Ours} & \textbf{74.74} & \textbf{58.19} & \textbf{26.88} & \textbf{53.27}
& \textbf{65.53} & \textbf{58.30} & \textbf{29.39} & \textbf{51.07}
& \textbf{79.10} & \textbf{55.24} & \textbf{11.93} & \textbf{48.76} \\
 & \textit{w/o Disc. Refine} & 73.02 & 56.81 & 25.54 & 51.79
& 63.94 & 55.94 & 28.91 & 49.60
& 77.28 & 53.27 & 8.60 & 46.38 \\
\multirow{-3}{*}{\centering 5\%} & \textit{w/o Sim. Penalty} & 74.08 & 56.47 & 25.96 & 52.17
& 63.55 & 57.61 & 27.88 & 49.68
& 78.82 & 54.83 & 10.64 & 48.10 \\
\bottomrule[1pt]
\end{tabular}
}
\vspace{-2mm}
\caption{Ablation study. Here, we use the ST model for AM.}
\vspace{-4mm}
\label{tab:ablation}
\end{table*}

\begin{figure*}[h]
  \centering
  \begin{subfigure}[h]{0.32\textwidth}
    \centering
    \includegraphics[width=\textwidth]{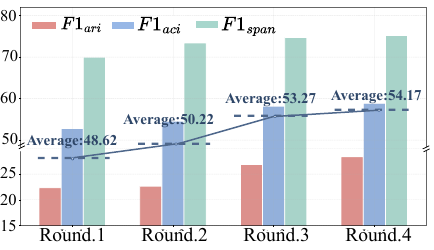}
    \caption{On the AAEC dataset.}
    \label{fig:loop1}
  \end{subfigure}%
  \hfill
  \begin{subfigure}[h]{0.32\textwidth}
    \centering
    \includegraphics[width=\textwidth]{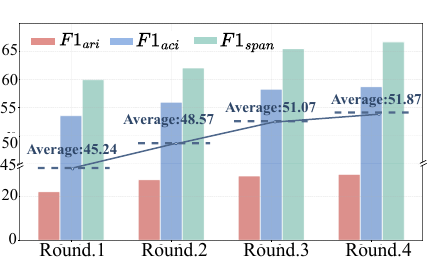}
    \caption{On the AbstRCT dataset.}
    \label{fig:loop2}
  \end{subfigure}%
  \hfill
  \begin{subfigure}[h]{0.32\textwidth}
    \centering
    \includegraphics[width=\textwidth]{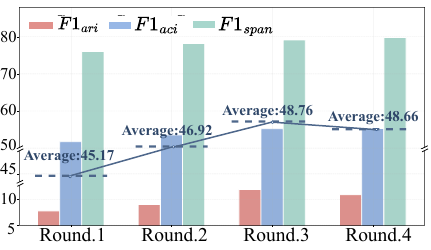}
    \caption{On the CDCP dataset.}
    \label{fig:loop3}
  \end{subfigure}
  \vspace{-3mm}
  \caption{Performance comparison of different rounds. Here, we use the ST model for AM.}
  \label{fig:loop}
  \vspace{-3mm}
\end{figure*}

\begin{figure*}[h]
\centering
  \includegraphics[width=0.95\textwidth]{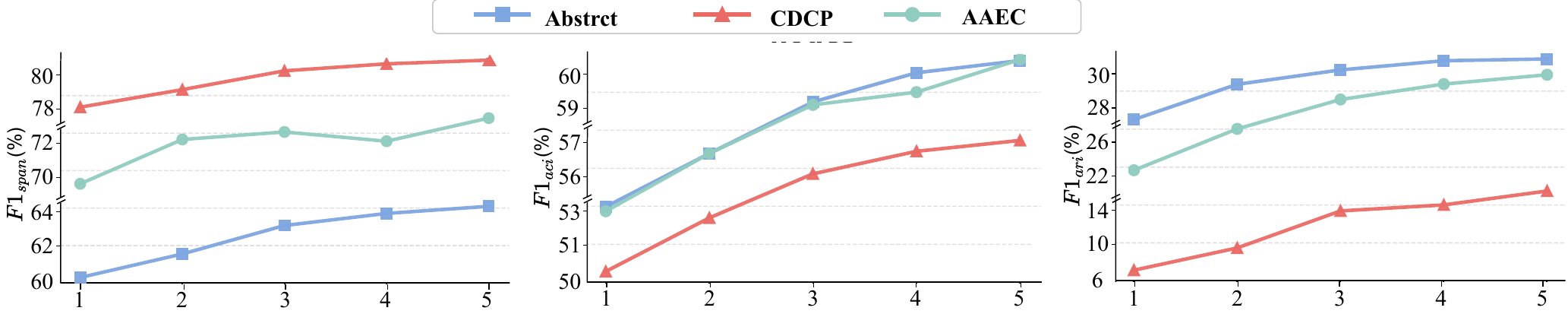}
  \vspace{-3mm}
  \caption{Performance comparison under different numbers of synthetic samples ($K$) using the ST model for AM.}
\vspace{-3mm}
\label{fig:K}
\end{figure*}

\subsection{Main Results}
Table ~\ref{tab:main result} reports the main results of our proposed framework and other synthetic data generation methods on the ST, UniASA, and Directly-SFT models. Here, “Origin” denotes training using only the original annotated data. From the results, we draw the following observations.

(1) Incorporating the synthetic data generated by our framework consistently improves the performance of all three AM models across all datasets and data settings.
Compared with training solely on the original data, our framework achieves stable gains under both full-data and low-resource scenarios, indicating that its generated data provides effective complementary training signals.
In particular, under the 5\%-setting, the ST model achieves average F1 gains of 8.06\%, 8.14\%, and 5.67\% on the three datasets, respectively. This shows that our framework can generate synthetic AM data with both high quality and strong diversity.

(2) For PLM-based AM models (ST and UniASA), the improvements brought by our framework are especially evident in the low-resource 5\%-setting, suggesting that the proposed framework effectively alleviates the problem of limited annotated data.
However, this trend becomes less pronounced for the LLM-based AM model (Directly-SFT). For example, on the AAEC dataset, our framework improves the SFT model by 4.47\% under the 5\%-setting and 5.16\% under the full-data setting, showing only a small difference between the two scenarios.
We conjecture that LLM-based AM models are inherently less sensitive to data scarcity due to their stronger prior knowledge.

(3) Our framework generally outperforms other synthetic data generation methods, including the GPT-4o-based QOS+DOS approach. Although these baselines also provide improvements over “Origin”, the gains achieved by our framework are more consistent and substantial across different datasets, AM models, and settings. For instance, compared with the QOS+DOS baseline, our framework achieves average F1 improvements of 2.22\%, 2.6\%, and 1.84\% on the three datasets under the 5\%-setting.
This indicates that the synthetic data generated by our framework is of higher quality and better suited for end-to-end AM tasks.

\subsection{Ablation Study}
To verify the effectiveness of each component in our framework, we examine two variants:
(1) "\textit{w/o Disc. Refine}", which removes discriminator refinement during adversarial training, and (2) "\textit{w/o Sim. Penalty}", which removes the similarity penalty from the final reward.
The ablation results are shown in Table~\ref{tab:ablation}. We make the following observations.
First, removing discriminator refinement consistently degrades performance. This suggests that, without iterative refinement, the discriminator fails to distinguish increasingly challenging synthetic samples generated in later training stages, thereby reducing the quality of the reward signals.
Second, removing the similarity penalty also leads to noticeable performance drops. This indicates that the penalty effectively discourages overly similar generations and alleviates the negative impact of repetitive synthetic texts on AM model training.

\begin{table*}[t!]
\centering
\resizebox{\linewidth}{!}{
\begin{tabular}{llcccccccccccc}
\toprule[1.5pt]
 &  & \multicolumn{4}{c}{\textbf{AAEC}} 
 & \multicolumn{4}{c}{\textbf{AbstRCT}} 
 & \multicolumn{4}{c}{\textbf{CDCP}} \\
\cmidrule[\heavyrulewidth](lr){3-6}
\cmidrule[\heavyrulewidth](lr){7-10}
\cmidrule[\heavyrulewidth](lr){11-14}
\multirow{-2}{*}{\centering\textbf{Setting}} 
& \multirow{-2}{*}{\centering\textbf{Method}}
& \textbf{$\mathrm{F1}_{span}$} & \textbf{$\mathrm{F1}_{aci}$} & \textbf{$\mathrm{F1}_{ari}$} & \textbf{Avg.}
& \textbf{$\mathrm{F1}_{span}$} & \textbf{$\mathrm{F1}_{aci}$} & \textbf{$\mathrm{F1}_{ari}$} & \textbf{Avg.}
& \textbf{$\mathrm{F1}_{span}$} & \textbf{$\mathrm{F1}_{aci}$} & \textbf{$\mathrm{F1}_{ari}$} & \textbf{Avg.} \\
\midrule
\multirow{2}{*}{\centering 10\%}
 & Origin & 69.94 & 52.12 & 17.53 & 46.53 & 58.36 & 51.65 & 23.78 & 44.60 & 77.95 & 51.04 & 5.92 & 44.97 \\
 & Ours 
& \textbf{76.54} & \textbf{59.42} & \textbf{24.73} & \textbf{53.56} & \textbf{67.11} & \textbf{59.30} & \textbf{31.23} & \textbf{52.55} & \textbf{81.00} & \textbf{56.19} & \textbf{13.67} & \textbf{50.29} \\
\midrule

\multirow{2}{*}{\centering 30\%}
 & Origin & 74.41 & 60.17 & 19.04 & 51.21 & 62.31 & 55.78 & 27.51 & 48.53 & 79.36 & 55.31 & 12.10 & 48.92 \\
 & Ours 
& \textbf{80.31} & \textbf{66.47} & \textbf{25.84} & \textbf{57.54} & \textbf{69.01} & \textbf{62.03} & \textbf{33.76} & \textbf{54.93} & \textbf{82.31} & \textbf{61.26} & \textbf{19.10} & \textbf{54.22} \\
\midrule

\multirow{2}{*}{\centering 50\%}
 & Origin & 79.41 & 67.30 & 36.48 & 61.06 & 65.57 & 59.61 & 31.27 & 52.15 & 79.76 & 59.87 & 15.43 & 51.69 \\
 & Ours 
& \textbf{84.11} & \textbf{72.20} & \textbf{42.28} & \textbf{66.20} & \textbf{71.37} & \textbf{65.01} & \textbf{36.72} & \textbf{57.70} & \textbf{83.06} & \textbf{65.27} & \textbf{26.53} & \textbf{58.29} \\
\midrule

\multirow{2}{*}{\centering 70\%}
 & Origin
& 81.93 & 73.70 & 51.49 & 69.04 & 67.59 & 63.19 & 35.31 & 55.36 & 81.56 & 61.72 & 20.74 & 54.67 \\
 & Ours 
& \textbf{86.08} & \textbf{77.95} & \textbf{56.74} & \textbf{73.59} & \textbf{72.89} & \textbf{67.94} & \textbf{39.36} & \textbf{60.06} & \textbf{84.11} & \textbf{68.47} & \textbf{35.84} & \textbf{62.81} \\
\bottomrule[1pt]
\end{tabular}
}
\vspace{-2mm}
\caption{Experimental results under different training data settings using the ST model for AM.}
\vspace{-2mm}
\label{tab:different_settings}
\end{table*}

\begin{figure}[t]
  \centering
  \includegraphics[width=0.40\textwidth]{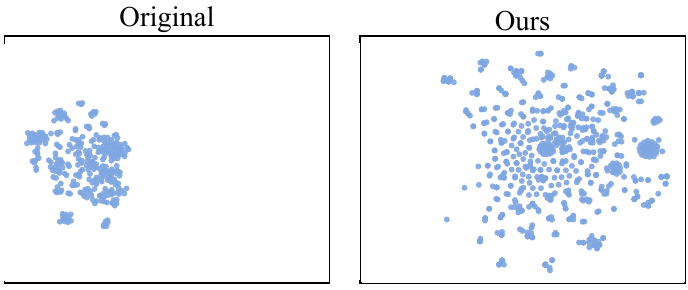}
  \vspace{-4mm}
  \caption*{(a) On the AAEC dataset.}
  \label{fig:SNE_AAEC}
  \includegraphics[width=0.40\textwidth]{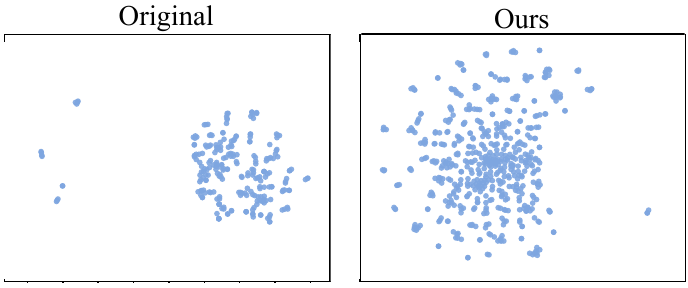}
  \vspace{-4mm}
  \caption*{(b) On the Abstrct dataset.}
  \label{fig:SNE_Abstrct}
  \includegraphics[width=0.40\textwidth]{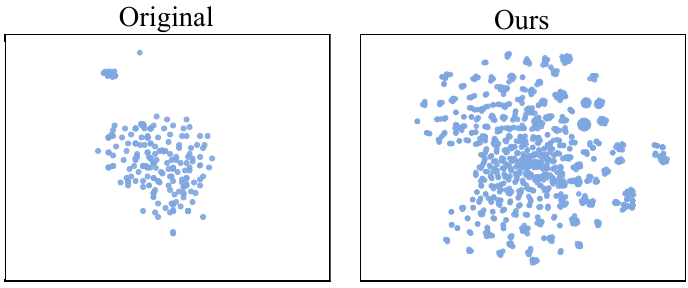}
  \vspace{-4mm}
  \caption*{(c) On the CDCP dataset.}
  \label{fig:SNE_cdcp}
  \caption{The t-SNE visualization of argument structure distributions under the full-data setting.}
  \label{fig:SNE}
  \vspace{-4mm}
\end{figure}

\subsection{Analysis and Discussion}
\paragraph{Effect of Evolution Rounds on Performance.}
We investigate how the rounds of evolution affect performance. Figure \ref{fig:loop} shows the results on the three datasets. Overall, as the number of evolution rounds increases, the average performance continues to improve, indicating that adversarial evolution can gradually enhance the quality of synthetic AM data. 
However, the improvements gradually plateau with more rounds, and slight fluctuations appear on CDCP, likely due to the optimization bottleneck between the generator and discriminator.
Considering the increasing training cost of iterative adversarial optimization, we set the number of evolution rounds to $r=3$ in the main experiments to balance effectiveness and efficiency.

\paragraph{Impact of the Number of Synthetic Samples.}
We further investigate the effect of the synthetic sample number $K$ on performance. Figure~\ref{fig:K} shows the performance trends under different $K$ values. Overall, all evaluation metrics consistently improve as $K$ increases. 
This suggests that introducing more synthetic candidate samples provides richer comparison signals for the discriminator, leading to more informative reward feedback for generator optimization. 
As a result, the generator can produce synthetic data with high-quality and diversity, which further benefits downstream AM models.

\paragraph{Performance under Different Training Data Settings.}
We further evaluate the performance gains of our framework under four training data settings (10\%, 30\%, 50\%, and 70\%). Table \ref{tab:different_settings} presents the relevant experimental results. 
Consistent with the main experimental results, our framework yields larger improvements in low-resource scenarios, while the gains gradually diminish as more original training data become available. 
An exception is observed for $\mathrm{F1}_{ari}$ on the CDCP dataset, which does not strictly follow this trend. We attribute this to the sparsity of argumentative relation labels in the original CDCP.

\paragraph{Argument Structural Diversity Analysis of Synthetic Data.} \label{Structural-Diversity}
To evaluate whether our framework improves the structural diversity of synthetic AM data, we follow \citeauthor{bao2025exploring} and visualize the distribution of argument structures. Specifically, we represent argument annotations as graphs, use graph2vec \citep{narayanan2017graph2vec} to obtain graph embeddings, and apply t-SNE for visualization. Notably, this process captures only structural information derived from argument component types and their relations, \textit{without incorporating textual semantic information}. As shown in Figure \ref{fig:SNE}, the synthetic data generated by our framework exhibit a broader and more dispersed structural distribution than the original data across all three datasets. These results indicate that our framework can not only generate structurally valid AM samples but also produce more diverse argument structures.

\paragraph{Comparison with Other Optimization Methods.}
To further compare our method with recent synthetic data generation approaches based on different optimization strategies, we select POPri and SEAL as additional baselines, where POPri represents preference-optimization-based data synthesis and SEAL represents reinforcement-learning-based data synthesis \citep{hou2025private,zweiger2026self}. Since neither method was originally designed for AM data synthesis, we adapt them to the task while preserving their core optimization mechanisms as much as possible, enabling them to generate complete text–structured argument annotation pairs\footnote{More detailed descriptions of POPri and SEAL, their task adaptation strategies, and the experimental settings are provided in Appendix \ref{appendix:method}}. Table \ref{tab:rl_pref_baselines} presents the experimental results under the 5\% low-resource setting using ST as the AM model. 
Both POPri and SEAL demonstrate a certain degree of effectiveness across the three datasets, indicating that general preference optimization and reinforcement learning methods can be transferred to AM data synthesis to some extent. However, our method achieves better overall performance than both approaches on all three datasets, further demonstrating the effectiveness of task-specific optimization for AM data synthesis.

\paragraph{Impact of Structure-Aware Reward.}
To further analyze the impact of explicit structural supervision on data synthesis, we introduce a structure-aware reward on top of the original reward function. For each generated text–annotation pair, we use an AM model trained only on the corresponding original training data and kept frozen as a verifier, and compute the structural reward based on the consistency between its predictions and the generated annotations in terms of argument component boundaries, component types, and argumentative relations, together with basic annotation validity checks. 
Table \ref{tab:structure_reward} presents the corresponding experimental results. We can see that introducing the structure-aware reward further improves the overall performance on AbstRCT and AAEC, indicating that explicit structural supervision can provide additional structural information beyond the existing adversarial feedback. In contrast, the overall performance on CDCP decreases slightly, possibly because the texts in this dataset are shorter and the argument structures are relatively simple, limiting the additional information provided by the structural reward. Overall, these results show that explicit structural supervision can serve as an effective complement to the adversarial reward in our framework, although its effect depends on the characteristics of the specific dataset.

In addition to the analyses above, we also provide supplementary discussions in the Appendix, including \nameref{Discourse-Structural-Diversity}, \nameref{Robustness-Analysis}, \nameref{Training Stability}, \nameref{Computational Efficiency} and \nameref{Case-Study}.

\section{Conclusion}
This paper proposes an \textit{adversarial reinforcement learning framework for data synthesis}, aimed at alleviating the scarcity of high-quality data in AM. The generator produces text-annotation pairs, while the discriminator provides quality feedback to guide iterative refinement. Experiments on three datasets show that our framework can significantly improve the performance of existing AM models and effectively alleviate the data scarcity problem.

\begin{table}[t]
\centering
\small
\resizebox{\linewidth}{!}{
\begin{tabular}{llcccc}
\toprule
\textbf{Dataset} 
& \textbf{Method} 
& \textbf{$\mathrm{F1}_{span}$} 
& \textbf{$\mathrm{F1}_{aci}$} 
& \textbf{$\mathrm{F1}_{ari}$} 
& \textbf{Avg.} \\
\midrule
\multirow{4}{*}{AAEC}
& POPri  & 72.64 & 54.91 & 21.62 & 49.72 \\
& SEAL   & 72.73 & 56.24 & 20.14 & 49.70 \\
& SFTSyn & 68.35 & 48.11 & 15.55 & 44.00 \\
& Ours   & \textbf{74.74} & \textbf{58.19} & \textbf{26.88} & \textbf{53.27} \\
\midrule
\multirow{4}{*}{AbstRCT}
& POPri  & 62.69 & 54.82 & 22.41 & 46.64 \\
& SEAL   & 63.84 & 55.86 & 20.10 & 46.60 \\
& SFTSyn & 57.14 & 49.93 & 19.71 & 42.26 \\
& Ours   & \textbf{65.53} & \textbf{58.30} & \textbf{29.39} & \textbf{51.07} \\
\midrule
\multirow{4}{*}{CDCP}
& POPri  & 76.88 & 51.89 & 9.10  & 45.96 \\
& SEAL   & 77.06 & 52.93 & 6.13  & 45.37 \\
& SFTSyn & 75.01 & 48.23 & 4.93  & 42.72 \\
& Ours   & \textbf{79.10} & \textbf{55.24} & \textbf{11.93} & \textbf{48.76} \\
\bottomrule
\end{tabular}
}
\caption{Performance comparison of different training methods using ST as the AM model.}
\label{tab:rl_pref_baselines}
\end{table}

\begin{table}[t]
\centering
\small
\setlength{\tabcolsep}{6pt}
\resizebox{\linewidth}{!}{
\begin{tabular}{llcccc}
\toprule
\textbf{Dataset} 
& \textbf{Method} 
& \textbf{$\mathrm{F1}_{span}$} 
& \textbf{$\mathrm{F1}_{aci}$} 
& \textbf{$\mathrm{F1}_{ari}$} 
& \textbf{Avg.} \\
\midrule
\multirow{2}{*}{AAEC}
& Ours                 & 74.74 & 58.19 & 26.88 & 53.27 \\
& \textit{w/} Structure Reward  & 75.24 & 59.93 & 27.07 & \textbf{54.08} \\
\midrule
\multirow{2}{*}{AbstRCT}
& Ours                 & 65.53 & 58.30 & 29.39 & 51.07 \\
& \textit{w/} Structure Reward  & 66.47 & 58.93 & 31.40 & \textbf{52.27} \\
\midrule
\multirow{2}{*}{CDCP}
& Ours                 & 79.10 & 55.24 & 11.93 & \textbf{48.76} \\
& \textit{w/} Structure Reward  & 77.95 & 54.14 & 12.64 & 48.24 \\
\bottomrule
\end{tabular}
}
\caption{Effect of the structure-aware reward on downstream AM performance using the ST model for AM.}
\label{tab:structure_reward}
\end{table}

\section*{Limitations}
Although the proposed framework achieves consistent improvements across multiple experimental settings, it still has several limitations.
\begin{itemize}
    \item First, to ensure the quality of the generated data, our method requires multiple rounds of optimization for both the generator and the discriminator, and the adversarial evolution process inevitably introduces additional computational overhead. Future work can further explore more efficient evolution strategies to reduce computational costs while maintaining the quality of synthetic data.

    \item Second, the effectiveness of our framework depends to some extent on the inherent capabilities of the generator and the discriminator. If the generator itself lacks stable long-text generation capability, the final performance will also be affected.

    \item Finally, our method still relies on some existing high-quality human-annotated data, and future work needs to explore methods for generating high-quality data from scratch.
\end{itemize}

\section*{Ethics Statement}
This paper uses large language models to generate synthetic data for the argument mining task. All experiments are based on publicly available AM datasets, and the generated samples are used only for data augmentation and model training. The chosen datasets are understood to be free of personally identifiable information and offensive content.

Since LLM-generated content may contain biases, factual errors, or invalid annotations, this paper performs format and structural validity checks on the generated samples to filter out low-quality data. However, automatic checking cannot fully guarantee semantic correctness, so human review is still recommended in practical applications or high-risk scenarios. During the writing of this paper, AI tools were used only for grammar checking and language polishing.

\section*{Acknowledgments}
This research was supported by the Natural Science Foundation for Top Talents of SZTU (Nos. GDRC202518 and GDRC202320).

\bibliography{References}

\newpage
\appendix
\section{Task Formulation} \label{appendix:Task-Formulation}
Given an input document $x$, the goal of end-to-end AM is to predict its complete argument structure $y=(Span, Comp, Rela)$, where $Span$ denotes the set of argument component spans, $Comp$ denotes the set of argument components with type labels, and $Rela$ denotes the set of relations between argument components. 

Specifically, this task consists of the following three parts:

(1) \textbf{Identifies the boundaries of argument components in $\textbf{x}$}. Each span is denoted as $s_i \in Span$, where $s_i=(b_i,e_i)$. Here, $b_i$ and $e_i$ denote the start and end character positions of the $i$-th argument component in $x$, respectively.

(2) \textbf{Assigns a corresponding component type to each span}. Each argument component is represented as $c_i=(b_i,e_i,l_i)$, where $l_i$ denotes the type label of the component, such as Claim or Evidence, and the specific label set is determined by the annotation scheme of each dataset.

(3) \textbf{Predicts the relations between pairs of argument components}. Each relation is represented as $r_{ij}=(b_i,e_i,b_j,e_j,l_{ij})$, where $(b_i,e_i)$ and $(b_j,e_j)$ denote the spans of the source and target components, respectively, and $l_{ij}$ denotes the relation label between them, such as Support or Attack.
Therefore, the overall objective of E2E-AM is to learn a mapping $f(x)\rightarrow(Span, Comp, Rela)$ that converts an input document into a structured argument representation.

\section{Algorithm} \label{appendix:Algorithm}
Below is the pseudo-code algorithm for the proposed framework.
\begin{algorithm}
\caption{Adversarial Evolution Algorithm}
\label{alg:adversarial_evolution}
\normalsize
\begin{spacing}{1.12}
\begin{algorithmic}[1]
\Statex \textbf{Input:} Original AM data $\mathcal{D}=\{z_i\}_{i=1}^{N}$;
initialization data of generator $\mathcal{D}_{G}^{0}$;
initialization data of discriminator $\mathcal{D}_{D}^{0}$;
a base LLM $\mathcal{M}$;
number of synthetic samples $K$;
number of evolution rounds $T$;
\Statex \textbf{Output:} Evolved generator $G_{\theta}^T$
    \State $G^0 \leftarrow \textsc{SFT}(\mathcal{M}, \mathcal{D}_{G}^{0})$
    \hfill (Eq.~\ref{eq:sft})
    \State $D^0 \leftarrow \textsc{SFT}(\mathcal{M}, \mathcal{D}_{D}^{0})$
    \hfill (Eq.~\ref{eq:disc_output})
    \For{$r = 0,1,\ldots,T-1$}
        \For{each original sample $z=(x,y) \in \mathcal{D}$}
            \State $u \leftarrow \textsc{Serialize}(z)$ \textcolor{gray}{//prompt template enclosure}
            \State $\{\hat{z}_{k}^{\,r}\}_{k=1}^{K} \leftarrow G^r(u)$ \textcolor{gray}{//synthesize $K$ data}
            \State $C^{r} \leftarrow 
            \textsc{Shuffle}(\{z,\hat{z}_{1}^{\,r},\ldots,\hat{z}_{K}^{\,r}\})$
            \State $y_{\mathrm{position}} \leftarrow \textsc{Position}(z, C^{r})$
            \State $\mathbf{p}^{r} \leftarrow D^r(C^{r})$
            \hfill (Eq.~\ref{eq:disc_output})
            
            \State $R_{\mathrm{dis}}^{r} \leftarrow
            \min\big(-\log(p_{\phi_r}(y_{\mathrm{position}}\mid C^{r})+\epsilon), c\big)$
            \hfill (Eq.~\ref{eq:disc_reward}) \textcolor{gray}{//adversarial reward}

            \For{$k = 1,2,\ldots,K$}
                \State $R_{\mathrm{sim}}^{k,r} \leftarrow 
                \textsc{SourcePenalty}(\hat{z}_{k}^{\,r}, z)$ \textcolor{gray}{//rule reward}
                \State $R_{k}^{r} \leftarrow 
                R_{\mathrm{dis}}^{r} + \lambda_1 \times R_{\mathrm{sim}}^{k,r}$
                \hfill (Eq.~\ref{eq:final_reward})
            \EndFor
        \EndFor
        \State $G^{r+1} \leftarrow  \textsc{GRPO}(G^r, \{R_{k}^{r}\})$
        \hfill (Eq.~\ref{eq:grpo_loss})

        \State $\hat{\mathcal{D}}^{\,r+1} \leftarrow G^{r+1}(\mathcal{D},K)$
        
        \State $\mathcal{D}_{D}^{r+1} \leftarrow 
        \textsc{BuildDiscData}(\mathcal{D},\hat{\mathcal{D}}^{\,r+1})$
        
        \State $D^{r+1} \leftarrow \textsc{SFT}(D^r, \mathcal{D}_{D}^{r+1})$
        \hfill (Eq.~\ref{eq:disc_output})
    \EndFor
    \State \Return $G^T$
\end{algorithmic}
\end{spacing}
\end{algorithm}

\section{Experimental Setting}
\subsection{Datasets} \label{appendix:datasets}
We conducted experiments on three widely used AM datasets: \textbf{AbstRCT} \citep{mayer2020transformer}, \textbf{AAEC} \citep{stab2017parsing}, and \textbf{CDCP} \citep{park2018corpus}. These datasets are derived from three distinct real-world argumentative scenarios and were manually annotated and constructed according to rigorous guidelines, making them high-quality gold-standard corpora in AM. All three datasets are English-language corpora.
Table \ref{tab:data_split} presents the statistics of the three datasets. The details are as follows:

\paragraph{AbstRCT} An argument mining dataset for the biomedical domain, with data collected from clinical trial abstracts in the PubMed database. It contains 500 abstracts, annotated with 3,279 argument components and 2,060 relations. The component types include MajorClaim, Claim, and Evidence, while the relation types include Support, Attack, and Partial-Attack. 

\paragraph{AAEC} An English persuasive essay dataset for argument structure analysis, with data collected from students’ persuasive essay writing tasks. After manual annotation, it contains 402 complete essays, with 6,089 argument components and 3,832 argumentative relations annotated. The component types include MajorClaim, Claim:For, Claim:Against, and Premise, while the relation types include Support and Attack. 
Following \citeauthor{bao2025exploring}, we uniformly represent Claim:For and Claim:Against as Claim in this work. This is mainly because LLMs are unable to distinguish the subtle differences between the two types of claims, which does not affect the annotation capability of AM models.

\paragraph{CDCP} An argument evaluability corpus for real-world user comments, with data collected from public comments on debt collection regulations issued by the U.S. Consumer Financial Protection Bureau. This dataset contains 731 user comments, annotated with 4,779 argument components and 1,353 relations. The component types include Value, Fact, Policy, Testimony, and Reference, while the relation types include Reason and Evidence.

\begin{table}[t]
\centering
\small
\setlength{\tabcolsep}{7pt}
\begin{tabular}{lcccc}
\toprule
\textbf{Dataset} 
& \textbf{5\% Train} 
& \textbf{100\% Train} 
& \textbf{Vali.} 
& \textbf{Test} \\
\midrule
AAEC    & 15 & 289 & 32 & 80 \\
AbstRCT & 18 & 350 & 50 & 100 \\
CDCP    & 27 & 530 & 58 & 150 \\
\bottomrule
\end{tabular}
\caption{
Statistics of the datasets. 
The 5\% and 100\% settings differ only in the size of the training split, while the validation and test sets remain unchanged.
}
\label{tab:data_split}
\end{table}

\begin{table*}[t!]
\centering
\resizebox{\linewidth}{!}{
\begin{tabular}{lcccccccccccc}
\toprule[1.5pt]
 & \multicolumn{4}{c}{\textbf{AAEC}} 
 & \multicolumn{4}{c}{\textbf{AbstRCT}} 
 & \multicolumn{4}{c}{\textbf{CDCP}} \\
\cmidrule[\heavyrulewidth](lr){2-5}
\cmidrule[\heavyrulewidth](lr){6-9}
\cmidrule[\heavyrulewidth](lr){10-13}
\multirow{-2}{*}{\textbf{Setting}}
& \textbf{$\mathrm{F1}_{span}$} & \textbf{$\mathrm{F1}_{aci}$} & \textbf{$\mathrm{F1}_{ari}$} & \textbf{Avg.}
& \textbf{$\mathrm{F1}_{span}$} & \textbf{$\mathrm{F1}_{aci}$} & \textbf{$\mathrm{F1}_{ari}$} & \textbf{Avg.}
& \textbf{$\mathrm{F1}_{span}$} & \textbf{$\mathrm{F1}_{aci}$} & \textbf{$\mathrm{F1}_{ari}$} & \textbf{Avg.} \\
\midrule
QOS+DOS  
& 72.10 & 56.02 & 23.45 & 50.52 
& 63.04 & 56.22 & 28.18 & 49.15 
& 77.61 & 54.80 & 7.87 & 46.76 \\
\hline
Qwen3-1.7B+Qwen3-8B  
& 70.78 & 53.64 & 21.05 & 48.49 
& 59.26 & 54.53 & 22.47 & 45.42 
& 75.16 & 51.62 & 4.73 & 43.84 \\
Qwen3-4B+Qwen3-8B 
& 73.16 & 56.73 & 25.37 & 51.75 
& 63.31 & 56.80 & 26.27 & 48.79 
& 77.79 & 53.11 & 7.90 & 46.27 \\
Qwen3-8B+Qwen3-1.7B 
& 72.54 & 56.74 & 24.96 & 51.41 
& 63.55 & 56.64 & 27.11 & 49.10 
& 78.05 & 53.86 & 6.82 & 46.24 \\
Qwen3-8B+Qwen3-4B 
& 73.89 & 58.01 & 26.13 & 52.68 
& 64.99 & 57.75 & \textbf{29.45} & 50.73 
& 78.57 & 54.21 & 8.23 & 47.00 \\
Qwen3-8B+Longformer
& 71.21 & 54.74 & 23.42 & 49.79
& 62.32 & 55.62 & 26.57 & 48.17
& 77.48 & 52.87 & 6.01 & 45.45 \\
Qwen3-8B+Qwen3-8B
& \textbf{74.74} & \textbf{58.19} & \textbf{26.88} & \textbf{53.27}
& \textbf{65.53} & \textbf{58.30} & 29.39 & \textbf{51.07}
& \textbf{79.10} & \textbf{55.24} & \textbf{11.93} & \textbf{48.76} \\
\bottomrule[1pt]
\end{tabular}
}
\caption{Robustness analysis under different generator--discriminator model combinations using ST as the AM model. Here, $\mathcal{M}_a$+$\mathcal{M}_b$ denotes using model $\mathcal{M}_a$ as the generator and model $\mathcal{M}_b$ as the discriminator. Longformer refers to Longformer-4096. In the main experiments, both the generator and discriminator are based on Qwen3-8B.}
\label{tab:model_scale_robustness}
\end{table*}

\begin{figure*}[t]
  \centering
  \begin{subfigure}[h]{0.32\textwidth}
    \centering
    \includegraphics[width=\textwidth]{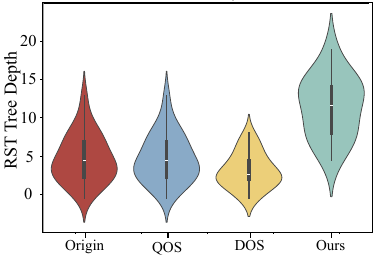}
    \caption{On the AAEC dataset.}
    \label{fig:RST_AAEC}
  \end{subfigure}%
  \hfill
  \begin{subfigure}[h]{0.32\textwidth}
    \centering
    \includegraphics[width=\textwidth]{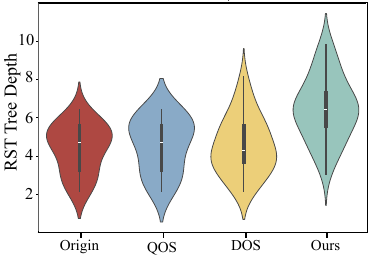}
    \caption{On the AbstRCT dataset.}
    \label{fig:RST_Abstrct}
  \end{subfigure}%
  \hfill
  \begin{subfigure}[h]{0.32\textwidth}
    \centering
    \includegraphics[width=\textwidth]{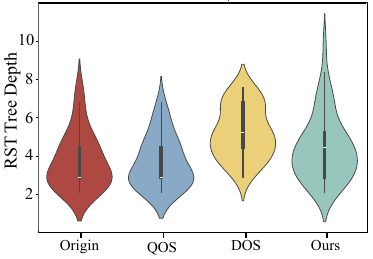}
    \caption{On the CDCP dataset.}
    \label{fig:RST_CDCP}
  \end{subfigure}
  \vspace{-3mm}
  \caption{RST tree depth distributions of different argument data under the full-data setting.}
  \label{fig:RST}
  \vspace{-4mm}
\end{figure*}

\subsection{Compared Method} \label{appendix:method}
We compare our data synthesis framework with the following data augmentation and synthetic data generation methods:
\begin{itemize}
    \item \textbf{Easy Data Augmentation} (EDA). This method only performs lexical-level operations on argumentative texts, while the original EDA framework includes lexical operations such as random insertion, deletion, replacement, and synonym replacement \citep{wei-zou-2019-eda}. However, since the first three operations may affect the argument structure of the original data, we only apply synonym replacement to the argumentative texts.
    \item \textbf{Few-shot Text Generation and Auto-annotation} (FTGA). FTGA first prompts GPT-4o with several original argumentative texts and asks it to generate new texts following the same domain and writing style. The LLM is not required to generate AM annotations in this setting. Instead, use the AM model trained on the original training set to automatically annotate the generated texts.
    \item \textbf{Joint Text and Label Synthesis} (JTLS). JTLS provides GPT-4o with a few text-annotation examples and requires it to generate both the argumentative text and the corresponding AM annotations.
    \item \textbf{Quality-Oriented Synthesis} (QOS) \citep{bao2025exploring}.QOS aims to produce high-quality synthetic samples by preserving the original argument structure. Specifically, it remains aligned with the original component and relation annotations, and only performs structure-aware paraphrasing on context and component. In this way, QOS emphasizes annotation reliability and structural consistency.
    \item \textbf{Diversity-Oriented Synthesis} (DOS) \citep{bao2025exploring}.DOS focuses on increasing the diversity of synthetic AM data. Instead of mainly rewriting the original instance, it encourages the generation of new samples with different topics, expressions, and argumentation patterns. 
    \item \textbf{Supervised Fine-Tuned Synthesis} (SFTSyn). SFTSyn is an additional LLM-based synthesis baseline. We construct paired training instances where the input is an original labeled AM sample and the output is a corresponding synthetic labeled sample. 
    LLM is then supervised and fine-tuned to learn this mapping and directly generate new text-annotation pairs.
    \item \textbf{Policy Optimization for Private Data} (POPri) \citep{hou2025private}. POPri is a preference optimization-based synthetic data generation method that scores and ranks multiple candidate samples to construct preference pairs and further optimizes the generator through DPO. To adapt it to AM data synthesis, we treat the complete text–annotation pair as the generation target and construct preference pairs based on candidate scores computed against the original AM training data. The privacy-related components in the original framework are omitted since they are not applicable to our setting.
    \item \textbf{Self-Adapting LLMs} (SEAL) \citep{zweiger2026self}. SEAL is a reinforcement learning-based method that optimizes synthetic finetuning data according to their utility for subsequent model updates and downstream performance. We adapt its self-edit generation process to produce complete AM text–annotation pairs. To reduce the computational cost of repeatedly performing temporary model updates, we use a lightweight proxy reward based on structural validity, similarity, and novelty to select self-edits for policy refinement.
\end{itemize}

For a fair comparison, POPri and SEAL use Qwen3-8B and start from the same generator SFT initialization as our method. They follow the same data splits and synthetic-data budget.

\section{Additional Analysis}

\subsection{Robustness Analysis under Different Model Configurations}\label{Robustness-Analysis}
In the above experiments, both the generator and discriminator are based on Qwen3-8B. To further evaluate the robustness of our framework, we investigate the effect of different generator–discriminator model-size combinations. Table~\ref{tab:model_scale_robustness} presents the results using ST as the AM model. 

The results show that the final performance is more sensitive to the scale of the generator than that of the discriminator. Specifically, performance drops noticeably under the Qwen3-1.7B+Qwen3-8B setting. In contrast, when the generator remains unchanged, replacing the discriminator with a smaller model leads to only marginal degradation. This indicates that the generator plays a more important role in determining the quality of synthetic AM data. 
Moreover, even under the Qwen3-4B+Qwen3-8B setting, our framework still achieves performance comparable to or better than the SOTA QOS+DOS baseline on most metrics, demonstrating the robustness of our framework across different model-size configurations.

In addition, we further examine the performance when using a lighter-weight model as the discriminator. Specifically, we use Longformer-Base-4096 as the discriminator while keeping the generator and all other experimental settings unchanged \citep{Beltagy2020Longformer}. We choose Longformer rather than BERT-base because it can handle longer input sequences and is therefore better suited for discriminating complete AM text–annotation candidate sets. As shown in Table \ref{tab:model_scale_robustness}, using Longformer as the discriminator leads to a decline in overall performance across all three datasets. This indicates that, compared with a lighter-weight encoder model, an LLM-based discriminator can provide more effective feedback signals to the generator.

\begin{table*}[t!]
\centering
\small
\resizebox{\linewidth}{!}{
\begin{tabular}{llcccccc}
\toprule
\textbf{Dataset} & \textbf{Stage} & \textbf{GRPO Reward} & \textbf{Reward Std.} & \textbf{Grad. Norm} & \textbf{Entropy} & \textbf{D Acc. (Beg.)} & \textbf{D Acc. (End.)} \\
\midrule
\multirow{4}{*}{AbstRCT}
& SFT     & -- & -- & -- & -- & 25.0\% & 69.1\% \\
& Round 1 & $1.72 \pm 0.057$ & $0.35 \pm 0.095$ & $0.08 \pm 0.044$ & $0.26 \pm 0.008$ & 55.0\% & 80.0\% \\
& Round 2 & $1.79 \pm 0.034$ & $0.25 \pm 0.101$ & $0.07 \pm 0.007$ & $0.25 \pm 0.007$ & 63.0\% & 86.2\% \\
& Round 3 & $1.82 \pm 0.032$ & $0.23 \pm 0.076$ & $0.07 \pm 0.020$ & $0.24 \pm 0.016$ & 67.0\% & 88.4\% \\
\midrule
\multirow{4}{*}{CDCP}
& SFT     & -- & -- & -- & -- & 25.0\% & 76.4\% \\
& Round 1 & $1.87 \pm 0.026$ & $0.13 \pm 0.017$ & $0.10 \pm 0.013$ & $0.26 \pm 0.008$ & 48.4\% & 81.2\% \\
& Round 2 & $1.88 \pm 0.029$ & $0.13 \pm 0.008$ & $0.12 \pm 0.032$ & $0.24 \pm 0.007$ & 58.1\% & 83.7\% \\
& Round 3 & $1.91 \pm 0.028$ & $0.14 \pm 0.014$ & $0.11 \pm 0.055$ & $0.22 \pm 0.008$ & 66.3\% & 85.6\% \\
\midrule
\multirow{4}{*}{AAEC}
& SFT     & -- & -- & -- & -- & 29.1\% & 82.0\% \\
& Round 1 & $1.72 \pm 0.057$ & $0.38 \pm 0.080$ & $0.07 \pm 0.003$ & $0.27 \pm 0.002$ & 50.0\% & 71.5\% \\
& Round 2 & $1.68 \pm 0.009$ & $0.38 \pm 0.065$ & $0.06 \pm 0.024$ & $0.25 \pm 0.004$ & 60.2\% & 90.0\% \\
& Round 3 & $1.69 \pm 0.020$ & $0.37 \pm 0.060$ & $0.06 \pm 0.020$ & $0.24 \pm 0.006$ & 64.3\% & 79.5\% \\
\bottomrule
\end{tabular}
}
\caption{Training dynamics of the generator and discriminator across adversarial evolution stages. GRPO Reward and Reward Std. respectively denote the reward obtained during generator optimization and the standard deviation of rewards. Grad. Norm and Entropy denote the gradient norm and policy entropy of the generator, respectively. D Acc. (Beg.) and D Acc. (End.) denote the discriminator accuracy at the beginning and end of each refinement stage. SFT denotes the discriminator initialization stage before adversarial evolution.}
\label{tab:training_dynamics}
\end{table*}

\begin{table}[t]
\centering
\small
\resizebox{\linewidth}{!}{
\begin{tabular}{llccc}
\toprule
\textbf{Dataset}
& \textbf{Methods}
& \textbf{Time}
& \textbf{Memory}
& \textbf{Avg.} \\
\midrule
\multirow{3}{*}{AbstRCT}
& Ours          & 9.3h & 60.3GB & \textbf{51.07} \\
& w/ Longformer & 8.6h & 51.1GB & 48.17 \\
& SFTSyn        & 0.9h & 43.4GB & 42.26 \\
\midrule
\multirow{3}{*}{CDCP}
& Ours          & 8.1h & 58.2GB & \textbf{48.76} \\
& w/ Longformer & 7.9h & 47.7GB & 45.45 \\
& SFTSyn        & 1.1h & 49.8GB & 42.72 \\
\midrule
\multirow{3}{*}{AAEC}
& Ours          & 7.6h & 64.1GB & \textbf{53.27} \\
& w/ Longformer & 7.4h & 54.7GB & 49.79 \\
& SFTSyn        & 0.8h & 46.7GB & 44.00 \\
\bottomrule
\end{tabular}
}
\caption{Computational cost and downstream AM performance of different synthesis methods. Time denotes the total training time in hours, Memory denotes the peak GPU memory usage in GB, and Avg. denotes the average of the three downstream AM F1 scores. We use ST as the AM model.}
\label{tab:efficiency_comparison}
\end{table}

\subsection{Discourse Diversity Analysis of Synthetic Data} \label{Discourse-Structural-Diversity}
To further analyze the diversity of the synthetic data generated by our framework at the discourse-structure level, we conduct a discourse structural analysis of argumentative content from different data sources. Specifically, we parse the data into Rhetorical Structure Theory (RST) trees and compare the distributions of their RST tree depths \citep{stede2016towards}. Unlike the previous \nameref{Structural-Diversity}, which does not introduce textual semantic information and constructs structural representations only from argument component types and their relations, this analysis directly operates on the generated texts themselves and characterizes the structural richness of synthetic data in terms of discourse organization and rhetorical relations. 

As shown in Figure \ref{fig:RST}, the data generated by our framework exhibit deeper RST tree structures across all three datasets. This indicates that our framework can generate synthetic texts with richer discourse organization structures, rather than merely producing shallow or linear argumentative expressions.
Notably, the RST tree-depth distribution of QOS is relatively close to that of the original data. This is because QOS mainly rewrites content while preserving the overall structural arrangement of the original samples, so its generated texts often retain discourse organization patterns similar to those of the original data.

\subsection{Adversarial Training Stability and Mode Collapse Analysis} \label{Training Stability}
To further analyze the training stability during adversarial evolution, we record the training dynamics of the Generator and Discriminator at different evolution stages, as shown in Table \ref{tab:training_dynamics}. For the Generator, the GRPO reward shows an overall mild upward trend, with only slight fluctuations on AAEC. Meanwhile, the reward standard deviation remains non-zero throughout training, indicating that informative reward differences are preserved among candidate samples. The gradient norm remains relatively small and stable without obvious abnormal fluctuations, while policy entropy decreases gradually rather than collapsing rapidly. Overall, these results indicate that the Generator optimization process is relatively stable, with no obvious reward collapse, gradient explosion, or premature policy collapse observed.

For the Discriminator, we evaluated its prediction accuracy before and after each incremental refinement round. We observe a consistent alternating adaptation pattern: after each Generator update, the Discriminator accuracy temporarily decreases because the newly generated synthetic samples become more challenging, but it consistently recovers and further improves after the subsequent refinement stage. This behavior demonstrates that the Generator continuously produces harder synthetic examples while the Discriminator successfully adapts to the evolving data distribution, leading to stable adversarial co-evolution rather than unstable competition.

This phenomenon also provides a reference for analyzing potential mode collapse. In AM data synthesis, mode collapse manifests as the Generator gradually concentrating on a small number of repetitive or highly similar argumentation patterns. If obvious mode collapse occurred, the Discriminator should be able to identify these repetitive patterns more easily, rather than repeatedly exhibiting accuracy drops after Generator updates. The current results instead show that newly generated samples in each round can continuously pose new challenges to the existing Discriminator. In addition, discriminator feedback encourages the Generator to produce samples that are more difficult to distinguish, while the similarity penalty further suppresses excessive reuse of the original samples. Therefore, within the adversarial evolution rounds examined, we do not observe obvious signs of mode collapse.

\subsection{Computational Efficiency Analysis}   \label{Computational Efficiency}
To assess the computational cost of our method, we compare different approaches in terms of training time, peak GPU memory usage, and downstream AM performance, as shown in Table \ref{tab:efficiency_comparison}. Compared with the variant using Longformer as the discriminator, our method introduces additional computational and GPU memory overhead, while achieving better downstream performance on all three datasets. In contrast, SFTSyn requires substantially less training time and computational resources, whereas our method additionally involves candidate sampling, GRPO optimization, and iterative discriminator updates, resulting in higher computational costs. However, these additional costs are accompanied by more substantial improvements in AM performance. It should be noted that the generated synthetic data can be reused to train different AM models without introducing additional overhead during downstream inference.

\subsection{Case Study} \label{Case-Study}
To further intuitively demonstrate the effectiveness of our framework, we select an AM sample for case analysis. We compare the original sample with the generated synthetic sample and examine both the generated text and its structured annotations. Specifically, we focus on 1) whether the text is natural and fluent, 2) whether the argument component spans can accurately match the input text, 3) whether the component labels are correct, and 4) whether the argumentative relations are consistent with the textual logic and semantics.

\begin{figure*}[t]
  \includegraphics[width=0.95\textwidth]{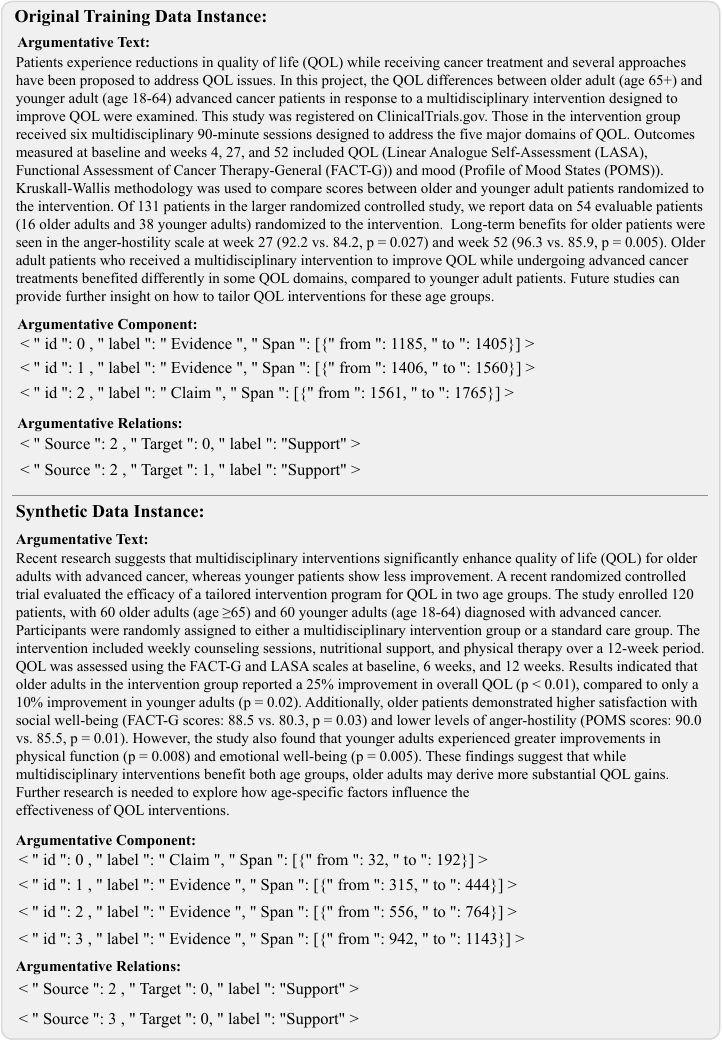}
  \caption{Examples of original data and synthetic data.}
\vspace{-2mm}
\label{fig:case}
\end{figure*}

As shown in Figure \ref{fig:case}, we require the model to generate a synthetic sample that is consistent with the topic discussed in the original sample. Even though some data and indicators from the original sample are used, the synthetic sample still exhibits clear differences in its specific content organization. Consequently, the annotatable argument components and logical relations also change accordingly.

\section{Prompt}
For clarity and reproducibility, we provide the prompt used in each step of our framework.

\begin{figure*}[t]
  \includegraphics[width=0.95\textwidth]{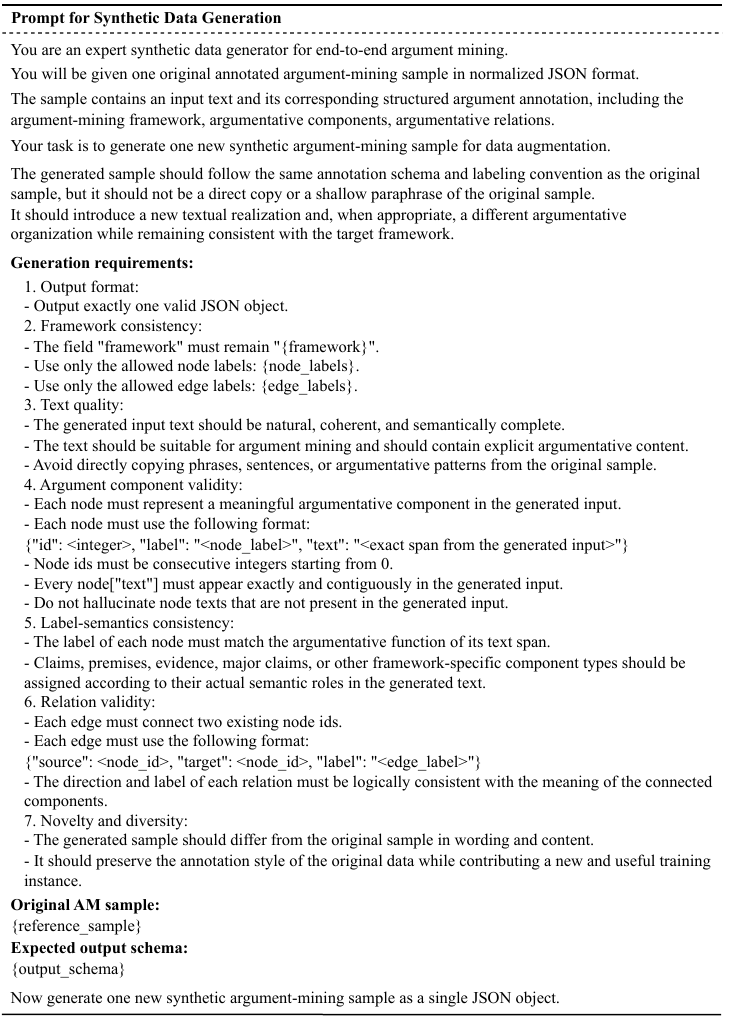}
  \caption{Prompt for Synthetic Data Generation.}
\vspace{-2mm}
\label{fig:gene_prompt}
\end{figure*}

\begin{figure*}[t]
  \includegraphics[width=0.95\textwidth]{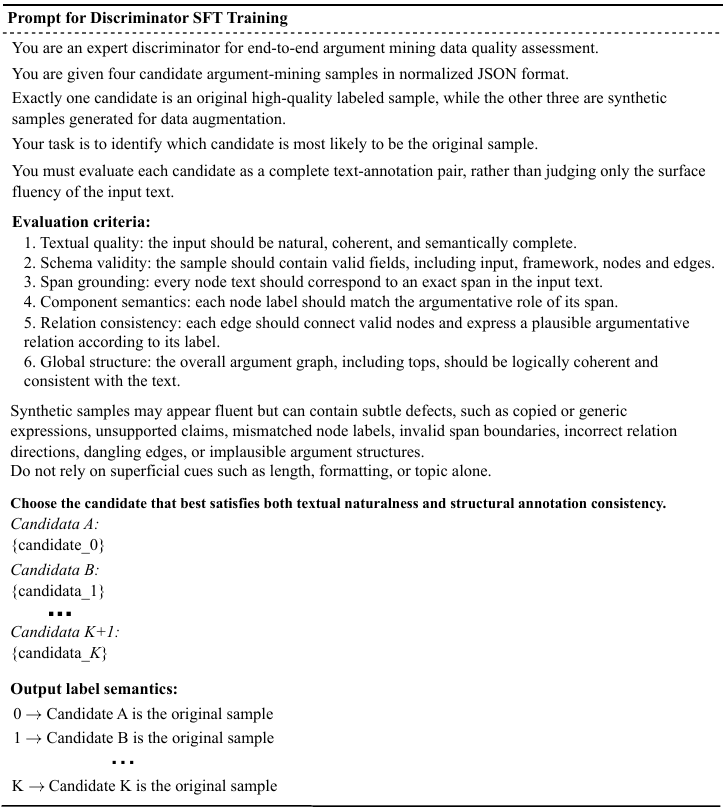}
  \caption{Prompt for Discriminator SFT Training.}
\vspace{-2mm}
\label{fig:disc_prompt}
\end{figure*}

Specifically, the prompt for generator initialization is shown in Figure \ref{fig:gene_prompt}. And the prompt for discriminator initialization is shown in Figure \ref{fig:disc_prompt}.

\end{document}